\documentclass[runningheads]{llncs}

\usepackage{eccv}

\usepackage{eccvabbrv}

\usepackage{graphicx}
\usepackage{booktabs}

\usepackage[accsupp]{axessibility}  

\usepackage{hyperref}

\usepackage{orcidlink}
\usepackage{multirow}
\usepackage{adjustbox}
\usepackage{color, colortbl}
\usepackage{amssymb}
\usepackage{pifont}
\usepackage{float}
\usepackage{wrapfig}
\usepackage{multido}
\usepackage{tabu}
\begin{document}

\title{Author Guidelines for ECCV Submission} 

\titlerunning{HERGen}

\author{Fuying Wang\inst{1}\orcidlink{0000-0002-3313-6479} \and
Shenghui Du\inst{1} \and
Lequan Yu\inst{1}\orcidlink{0000-0002-9315-6527}}

\authorrunning{F.Wang, S.Du, and L.Yu}

\institute{The University of Hong Kong, Hong Kong SAR\\
\email{\{fuyingw@connect.,shenghui@connect.,lqyu@\}hku.hk}}

\definecolor{darkgreen}{rgb}{0,0.5,0}
\newcommand{\highlight}[1]{\textbf{\textcolor{darkgreen}{#1}}}

\newcommand{\para}[1]{\vspace{.05in}\noindent\textbf{#1}}
\newcommand{\modelname}{HERGen}
\title{\modelname: Elevating Radiology Report Generation with Longitudinal Data}

\maketitle

\begin{abstract}

Radiology reports provide detailed descriptions of medical imaging integrated with patients' medical histories, while report writing is traditionally labor-intensive, increasing radiologists' workload and the risk of diagnostic errors.
Recent efforts in automating this process seek to mitigate these issues by enhancing accuracy and clinical efficiency. 
%
However, existing automated approaches are based on a single timestamp and often neglect the critical temporal aspect of patients' imaging histories, which is essential for accurate longitudinal analysis. 
To address this gap, we propose a novel \textbf{H}istory \textbf{E}nhanced Radiology \textbf{R}eport \textbf{Gen}eration (\textbf{\modelname}) framework that employs a group causal transformer to efficiently integrate longitudinal data across patient visits.
Our approach not only allows for comprehensive analysis of varied historical data but also improves the quality of generated reports through an auxiliary contrastive objective that aligns image sequences with their corresponding reports.
More importantly, we introduce a curriculum learning-based strategy to adeptly handle the inherent complexity of longitudinal radiology data and thus stabilize the optimization of our framework.
The extensive evaluations across three datasets demonstrate that our framework surpasses existing methods in generating accurate radiology reports and effectively predicting disease progression from medical images.


\end{abstract}
\section{Introduction}
\label{sec:intro}

\begin{figure}
    \centering
    \includegraphics[width=0.8\linewidth]{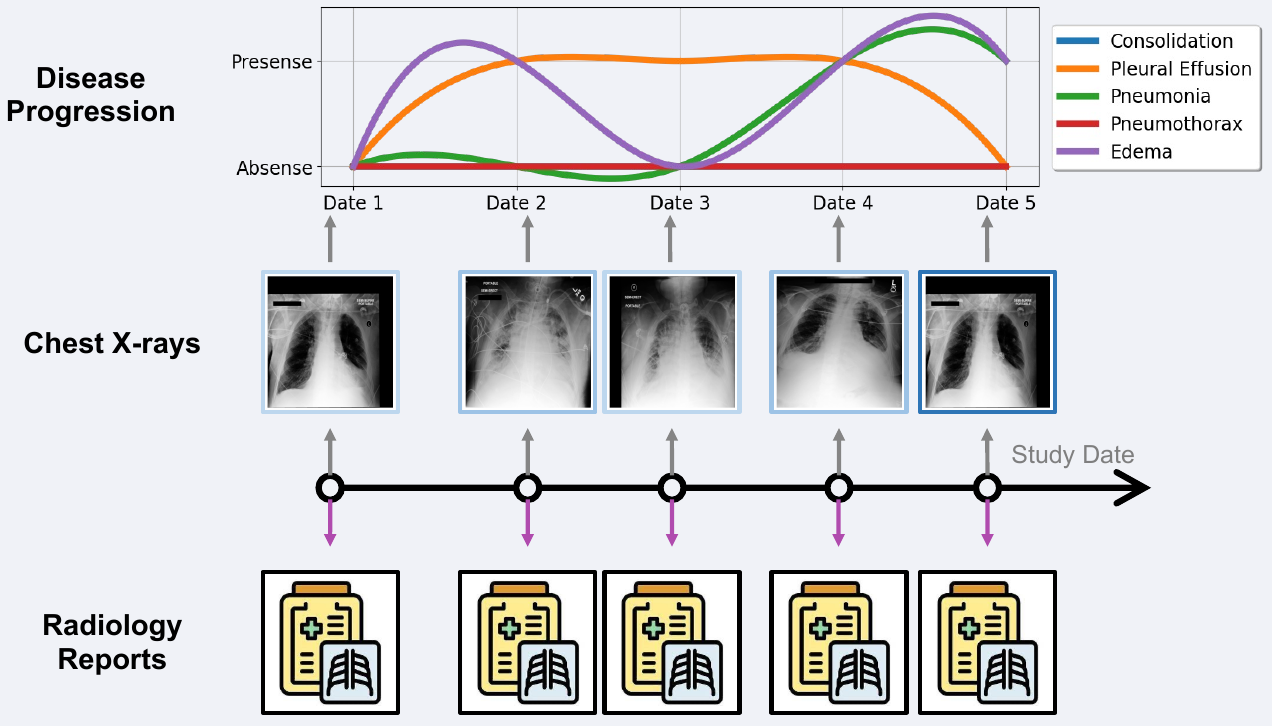}
    \caption{
    Overview of our \modelname \ for radiology report generation: Our model processes longitudinal data for each patient and utilizes the comprehensive historical information within these longitudinal data to generate robust and precise radiology reports. 
    }
    \label{fig:teaser}
\end{figure}

Chest X-rays are a cornerstone in diagnosing thoracic conditions, including pneumonia and lung cancer~\cite{johnson2019mimic, raoof2012interpretation}.
Given a chest X-ray, radiologists will meticulously examine each anatomical section in the X-ray and document their observations with detailed text descriptions.
The generated report is crucial to diagnose diseases (\eg, lung cancer, scoliosis) and assess the position of the treatment devices (\eg, tracheostomy tubes, pacemakers).
Particularly, when prior images are available, radiologists commonly compare the clinical findings of the current scan with prior scans to assess the evolution of disease over time, which is essential in regular clinical evaluations.
However, the high volume of chest X-rays overwhelms radiologists, exacerbating the impact of the global shortfall in this workforce~\cite{cao2023current, rimmer2017radiologist}.

Automated chest X-ray report generation has emerged as a key research area, aiming to ease radiologists' workload and improve patient care~\cite{thrall2018artificial}. 
Mainstream approaches focus on improving clinical accuracy and completeness of individual reports~\cite{chen2020generating, chen2022cross, nicolson2023improving, tanida2023interactive}, often overlooking the chronological consistency in longitudinal imaging.
Modeling such inherent temporal information in chest X-rays has shown to be crucial for generating precise radiology reports~\cite{karwande2022chexrelnet, bannur2023learning, zhu2023utilizing, serra2023controllable}. 
Some recent studies integrate prior images for temporal representation and enhance report generation~\cite{bannur2023learning, serra2023controllable}. However, they are limited to the use of only one prior image for the current report, failing to capture high-level disease progression evident across a patient's history.
This highlights the need for a framework that learns accurate representations from both study-level and patient-level images, thereby producing reports closely aligned with radiologists' analyses.

%
%

In this paper, we propose a novel \textbf{H}istory \textbf{E}nhanced radiology \textbf{R}eport \textbf{Gen}er-
ation framework (\modelname) to effectively capture the temporal information of longitudinal data for generating comprehensive and temporally coherent radiology reports, as shown in Fig.~\ref{fig:teaser}. 
The key part is a causal transformer model, which treats all visual tokens from the same image as a group and uses a group causal attention mechanism to handle it. 
Viewing all visual tokens of each patient as a sequence, this mechanism groups visual tokens from the same image, facilitating intra-image interactions of visual tokens and inter-image interactions of tokens only across previous studies. Notably, it treats each patient's X-ray series as a distinct sequence, adeptly handling the variability in the number of longitudinal images per patient. 
%
Moreover, we further refine the model's capability to chart disease progression through a cross-modal contrastive learning objective, ensuring the alignment of longitudinal visual representations with their narrative reports.
Due to the inherent complexity of longitudinal data, it is non-trivial to optimize the whole framework. 
We thereby introduce a new curriculum learning-based optimization strategy in three progressive steps to enhance and stabilize the learning process of our framework.
The model is first trained to generate radiology reports for individual images and then, we employ the auxiliary contrastive alignment module to optimize the latent space. After that, the entire framework is trained with the integration of a temporal aggregation module, enabling it to learn from the patient's historical information.
%
Extensive experimental results on radiology report generation and temporal medical image classification tasks demonstrate the superiority of our framework in generating accurate radiology reports and effectively predicting the disease progression from medical images.
The source code is available at \url{https://github.com/fuying-wang/HERGen}.
\section{Related Work}
\label{sec:related}

\para{Automated Report Generation.}
Radiology report generation, inspired by image captioning techniques~\cite{cornia2020meshed, you2016image, xu2015show, vinyals2015show}, face unique challenges due to the complexity and variability in radiology reports~\cite{alfarghaly2021automated, chen2022cross, chen2020generating, jing2017automatic, ma2021contrastive, liu2021exploring, wang2018tienet, you2021aligntransformer}.
Initial approaches, primarily based on CNN-RNN~\cite{jing2020show, jing2017automatic, wang2018tienet, zhang2020radiology}, have evolved with the adoption of transformer~\cite{vaswani2017attention}. 
Recent advancements include memory-driven transformers for enhanced cross-modal interactions~\cite{chen2020generating, chen2022cross}, alignment of visual features with disease tags~\cite{you2021aligntransformer}, and contrastive methods for anomaly detection~\cite{ma2021contrastive}.
Integration of knowledge graphs~\cite{liu2021exploring, zhang2020radiology}, warm starting strategies~\cite{nicolson2023improving}, and interactive frameworks for region-specific reports \cite{tanida2023interactive} have also been explored. 
However, these methods often treat X-rays and reports as independent entities, overlooking the temporal aspects inherent in various radiology modalities.

\para{Longitudinal Chest X-ray Representations.}
Radiology studies, inherently chronological, are crucial for accurate reporting, yet the temporal dimension is often under-addressed in research. \cite{ramesh2022improving} indirectly acknowledged the importance of sequential context by proposing a method to reduce language model hallucinations. 
\cite{bannur2023learning} introduced a self-supervised framework capturing the longitudinal evolution of chest X-ray findings. 
Similarly, \cite{zhu2023utilizing} developed a cross-attention-based multi-modal fusion framework utilizing patient record chronology to enhance report pre-filling. 
\cite{karwande2022chexrelnet} employed graph attention networks~\cite{velivckovic2017graph} for an anatomy-aware approach to tracking disease progression in longitudinal CXR data.
\cite{serra2023controllable} used Faster R-CNN~\cite{ren2015faster} to project longitudinal studies into a composite representation highlighting anatomical changes over time. 
However, most of these methods primarily focus on learning representations rather than generating reports.
Furthermore, these methods often treat two consecutive image-text pairs, lacking flexibility for varying patient history lengths and are limited in capturing the complex progression of diseases.

\para{Biomedical Vision-language Pretraining.}
Radiology reports, paired with chest X-rays, offer rich labels for learning visual representations.
Building on the CLIP framework~\cite{radford2021learning}, ~\cite{zhang2022contrastive, huang2021gloria, boecking2022making, wang2022multi} demonstrate the efficacy of self-supervised vision-language pretraining in biomedical imaging tasks.
Particularly, \cite{zhang2022contrastive} use a contrastive objective~\cite{oord2018representation} for modality alignment, \cite{huang2021gloria} focus on local alignment for detailed feature learning, and~\cite{boecking2022making} develop CXR-BERT, employing masked language modeling for enhanced image feature learning from radiology language. 
Furthermore, \cite{bannur2023learning} adapt BioViL~\cite{boecking2022making} for longitudinal analysis in radiology, improving temporal aspects in report generation and classification tasks. 
Our work further explores the application of vision-language pretraining on longitudinal data, aiming to effectively capture disease progression in patient records.
\section{Method}
\label{sec:method}


\subsection{Problem Formulation}
\label{sec: problem_formulation}

\begin{figure}[t]
    \centering
    \includegraphics[width=\linewidth]{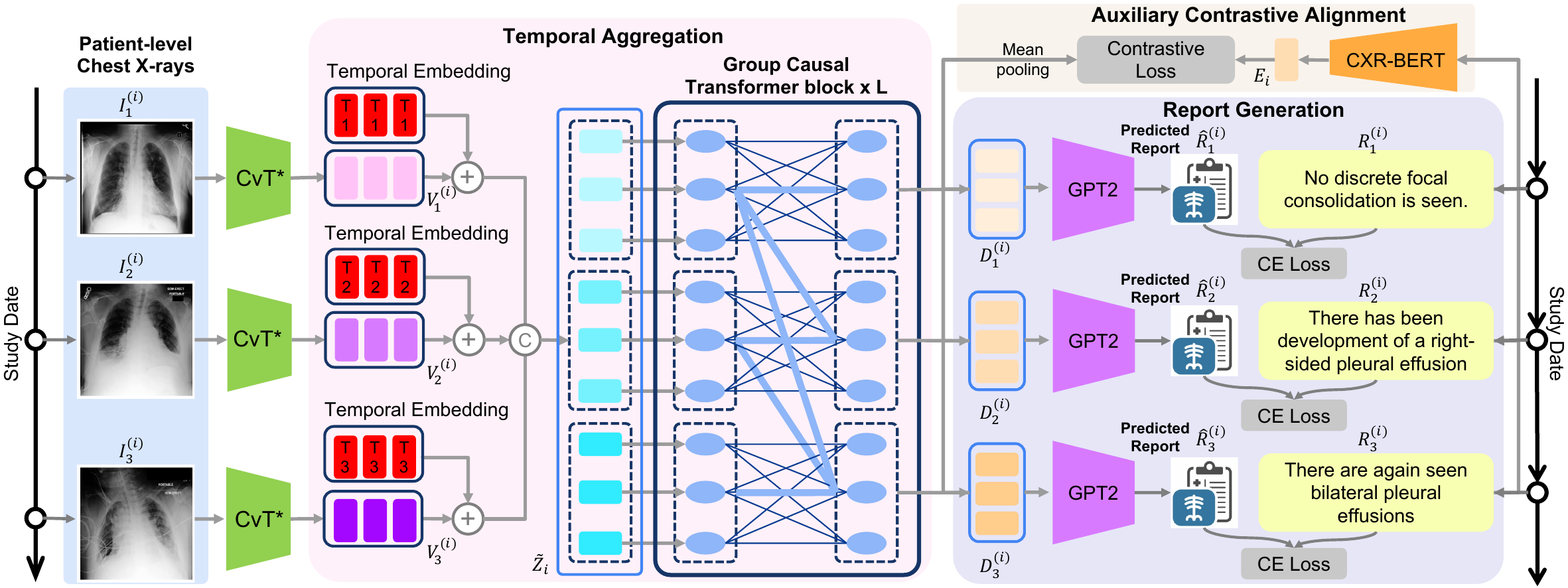}
    \caption{
    \textbf{H}istory \textbf{E}nhanced Radiology \textbf{R}eport \textbf{Gen}eration (\modelname): the framework processes patient-level chest X-rays using the CvT* (CvT combined with the encoder projection layer), which then aggregates temporal information through a group causal transformer. 
    Subsequently, GPT2 serves as the decoder for predicting the radiology report, which was optimized by a cross-entropy (CE) loss.
    Additionally, an auxiliary contrastive alignment module is employed to enhance the alignment of the latent spaces between image and text modalities, thereby producing more consistent reports.
    Note that in the group causal transformer block, thick lines represent image-level interactions, while thin lines indicate token-level interactions.
    }
    \label{fig:framework}
\end{figure}

The overall framework of the proposed method is shown in Fig.~\ref{fig:framework}.
We analyze a dataset comprising chest X-rays from $M$ patients, denoted as $\{\mathcal{I}_i \}_{i = 1, 2, ..., M}$, where $\mathcal{I}_i = 
\{\mathbf{I}_j^{(i)}\}_{j=1, 2, ..., N_i}$ represents the set of X-rays for the $i$-th patient and $N_i$ is the number of studies (visits).
For each patient $i$, their X-rays are chronologically ordered based on their associated study dates $\mathcal{T}_i = \{ \mathbf{T}_j^{(i)} \}_{j=1, 2, ..., N_i}$.
The objective of our method is to generate a set of radiology reports $\{ \mathcal{\hat{R}}_i \}_{i=1, 2, ..., M}$ for each patient, aiming to closely approximate the ground truth reports $\{ \mathcal{R}_i \}_{i=1, 2, ..., M}$.
In the following, we use $i$ and $j$ to index patient and study respectively.

\subsection{History-enhanced Report Generation}
\label{sec:hergen}

\para{Extract Representations of Single Images.}
In our approach, each X-ray image $\mathbf{I}_j^{(i)} \in \mathbb{R}^{C\times W \times H}$ is first encoded into a feature representation $\mathbf{P}_j^{(i)} \in \mathbb{R}^{S\times F}$ with an image encoder.
Here, $C$, $W$, and $H$ denote the number of channels, width, and height of the image, respectively, while $S$ and $F$ represent the number of visual tokens and the feature dimension per token.
Following CvT-212DistilGPT2~\cite{nicolson2023improving}, we utilize the CvT architecture~\cite{wu2021cvt}, pretrained on ImageNet-21K, as our image encoder, while our framework can take other encoder backbones. 
To tailor the dimensions of $S$ and $F$ to our requirements, we introduce an encoder projection layer $E_{proj}$. 
This layer comprises a $1\times 1$ convolution layer followed by a linear projection layer, transforming each $\mathbf{P}_j^{(i)}$ into a more compact visual representation $\mathbf{V}_j^{(i)} \in \mathbb{R}^{S' \times F'}$, where $S'$ and $F'$ denote the adjusted number of visual tokens and their new dimensionality, respectively.

\para{Sequential Date-aware Temporal Embedding.}
Temporal embeddings are especially critical for our group causal transformer to learn longitudinal information.
Standard positional embeddings typically assume equidistant intervals between tokens, an assumption that is not applicable in our context due to the varying time gaps between consecutive chest X-rays.
For example, the clinical progression captured in X-rays taken a month apart is significantly different from that in X-rays taken a year apart.
To tackle this challenge, we introduce study date-aware positional embeddings, $\mathbf{p}_j^{(i)} \in \mathbb{R}^{S^* \times F'}$ for each study.
%
These embeddings are conditioned on the study dates $\mathbf{T}_j^{(i)}$, offering a more precise representation of the temporal intervals between X-rays. 
In detail, we first calculate the relative study date for each X-ray image as $\mathbf{T'}_j^{(i)} = \mathbf{T}_j^{(i)} - \mathbf{T}_0^{(i)}$.
Then, we identify the maximum relative study date in the training set and create a learnable embedding vocabulary of the corresponding length. 
Each temporal embedding is defined as: $\mathbf{p}_j^{(i)} = \mathrm{Embedding}(\mathbf{T'}_j^{(i)}) \in \mathbb{R}^{1 \times F'}$.
The visual token embeddings $\mathbf{V}_j^{(i)}$ are then added with the temporal embedding $\mathbf{p}_j^{(i)}$ to form $\mathbf{Z}_j^{(i)} = \mathbf{V}_j^{(i)} + \mathbf{p}_j^{(i)}$, where $\mathbf{Z}_j^{(i)} \in \mathbb{R}^{S' \times F'}$.
Finally, we concatenate all visual token embeddings for each patient to create a patient-level sequence $\mathbf{\tilde{Z}}_i = \mathrm{Concat}([\{ \mathbf{Z}_j^{(i)} \}_{j=1, 2, ..., N_i} ])$ where $\mathbf{\tilde{Z}}_i \in \mathbb{R}^{S^{(i)} \times F'}$ and $S^{(i)} = N_i \times S'$, which is then fed into the group causal transformer for temporal aggregation.

\para{Group Causal Transformer.}
Our group causal transformer comprises $L$ group causal blocks, designed to aggregate longitudinal information from patient data. 
%
%
In block $l$, for every visual token (indexed by $p$), we first compute the query, key, and value vectors from its preceding block's representation $\mathbf{z}^{(l-1)}_{(p)} \in \mathbb{R}^{F'}$ as:
\begin{align}
    &\mathbf{q}_{(p)}^{(l, a)} = \mathbf{W}_Q^{(l, a)} \mathrm{LN}(\mathbf{z}_{(p)}^{(l-1)}) \in \mathbb{R}^{D_h}, \nonumber \\
    &\mathbf{k}_{(p)}^{(l, a)} = \mathbf{W}_K^{(l, a)} \mathrm{LN}(\mathbf{z}_{(p)}^{(l-1)}) \in \mathbb{R}^{D_h}, \nonumber \\
    &\mathbf{v}_{(p)}^{(l, a)} = \mathbf{W}_V^{(l, a)} \mathrm{LN}(\mathbf{z}_{(p)}^{(l-1)}) \in \mathbb{R}^{D_h}.
\end{align}

\begin{figure}[h]
    \centering
    \begin{minipage}[t]{0.48\textwidth}
        \includegraphics[width=\linewidth]{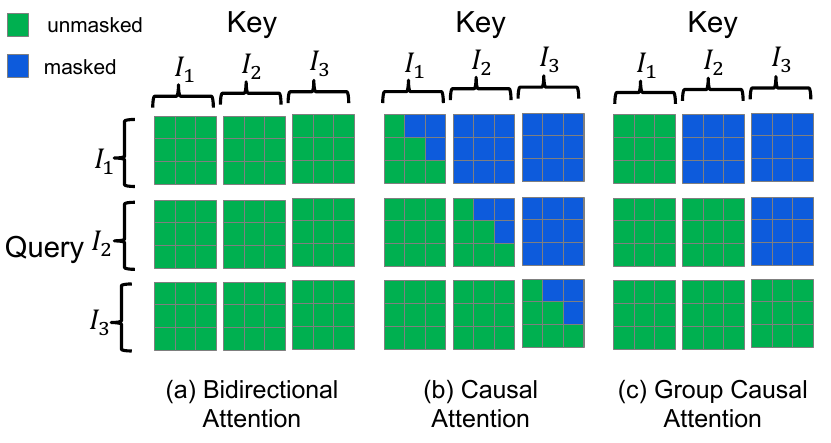}
        \centering
        \caption{Comparison between (a) bidirectional attention, (b) causal attention and (c) our group causal attention.}
        \label{fig:group causal attention}
    \end{minipage}
    \hfill
    \begin{minipage}[t]{0.48\textwidth}
        \includegraphics[width=\linewidth]{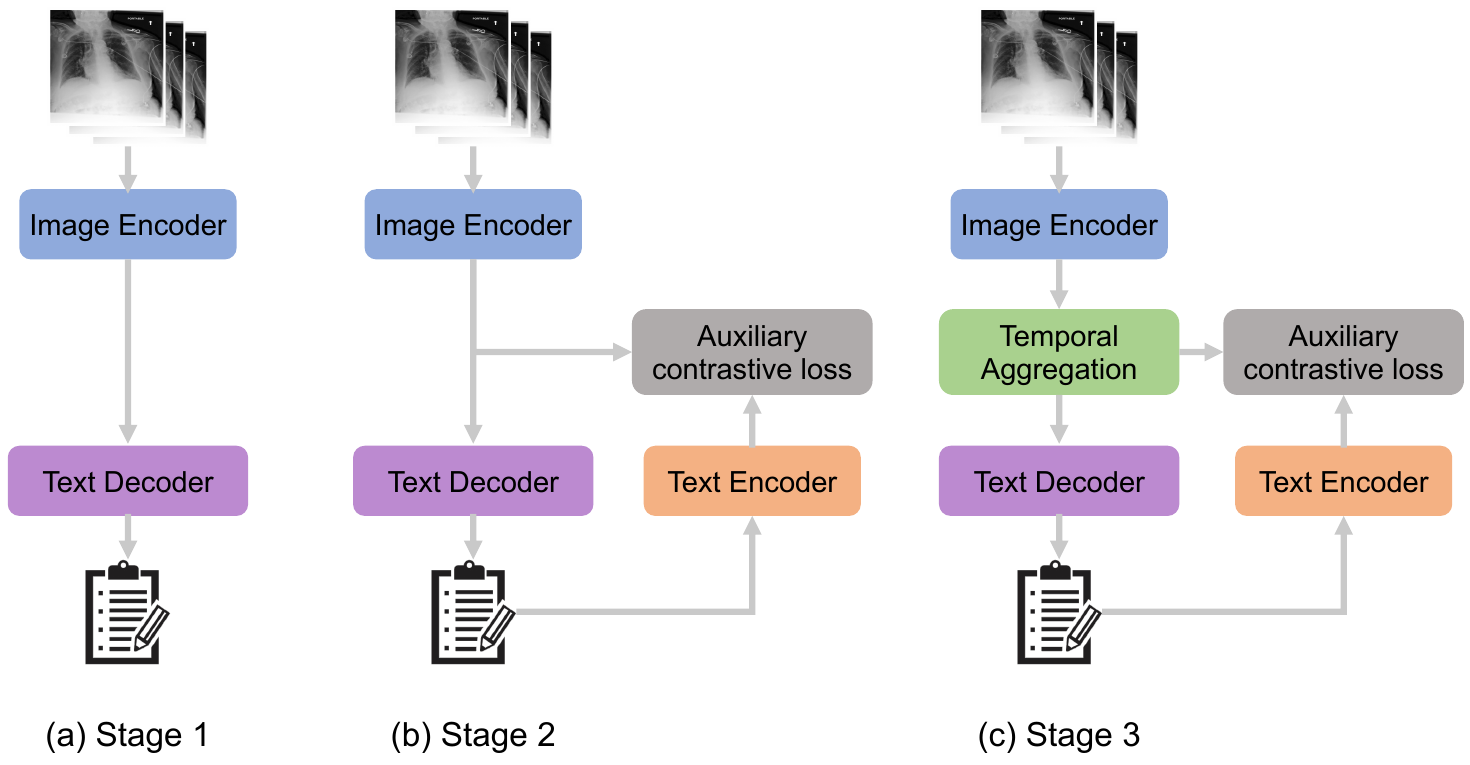}
        \centering
        \caption{Illustration of the proposed curriculum training strategy.}
        \label{fig:curriculum learning}
    \end{minipage}
\end{figure}

Here, $\mathrm{LN}$ denotes LayerNorm and $a=1, ..., A$ indexes the $A$ attention heads with the latent dimensionality for each head being $D_h = F' / A$.
The initial representation $\mathbf{z}^{(0)}$ corresponds to the input sequence $\mathbf{\tilde{Z}}_i$.
$\mathbf{W}_Q$, $\mathbf{W}_K$, and $\mathbf{W}_V$ are learnable matrices.
For simplicity, we have omitted the patient index $i$ in our notation.
Then, the process for computing dot-product self-attention weights, along with subsequent steps, is defined as follows:
\begin{align}
    \mathbf{\alpha}_{(p)}^{(l, a)} = \mathrm{SM}(\frac{\mathbf{q}_{(p)}^{(l, a)}}{D_h} \cdot [\mathbf{k}_{(p')}^{(l, a)}]_{p' \in 0, ..., S^{(i)} - 1}), \nonumber \\
    \mathbf{s}{(p)}^{(l, a)} = \sum_{p'} \mathbf{M}{(p')} \mathbf{\alpha}{(p)(p')}^{(l, a)} \mathbf{v}{(p')}^{(l, a)},
\end{align}
where $\mathbf{\alpha} \in \mathbb{R}^{S^{(i)}}$.
The group causal attention matrix $\mathbf{M}(p')$, as illustrated in Fig.~\ref{fig:group causal attention}, differs fundamentally from the bidirectional attention used in BERT~\cite{devlin2018bert} and the causal attention in GPT~\cite{radford2019language}.
It ensures that each visual token within an image not only interacts with others in the same image but also with tokens from preceding images. 

This design reflects our intention to make the transformer cognizant of the temporal sequence in radiological data, a crucial aspect for accurately capturing disease progression over time.
Subsequently, we perform concatenation followed by a Multi-Layer Perceptron (MLP) with residual connections to get the output.
This can be mathematically represented as:
\begin{align}
\mathbf{z'}_{(p)}^{(l)} &= \mathbf{W}_O [\mathbf{s}_{(p)}^{(l, 1)} ... \mathbf{s}_{(p)}^{(l, A)}] + \mathbf{z}_{(p)}^{(l-1)}, \nonumber \\
\mathbf{z}_{(p)}^{(l)} &= \mathrm{MLP}(\mathrm{LN} (\mathbf{z'}_{(p)}^{(l)})) + \mathbf{z'}_{(p)}^{(l)},
\end{align}
where $\mathbf{W}_O$ is a learnable matrix.
Then, the output sequence of the group causal transformer, denoted as $\mathbf{z}^{(L)} \in \mathbb{R}^{S^{(i)} \times F'}$, is split into a series of representations of studies $\{ \mathbf{D}_j^{(i)} \}_{j=1, ..., N_i}$ and $\mathbf{D}_j^{(i)} \in \mathbb{R}^{S' \times F'}$.

\para{Report Generation and Auxiliary Contrastive Alignment.}
Each temporally aggregated visual representation $\mathbf{D}_j^{(i)}$ is input into a text decoding module 
for generating radiology reports.
Note that We chose GPT-2 as the text decoder following~\cite{nicolson2023improving}, which shows DistilGPT2~\cite{sanh2019distilbert} outperforms other alternatives like ClinicalBERT~\cite{alsentzer2019publicly}, PubMedBERT~\cite{gu2021domain}, and SciBERT~\cite{beltagy2019scibert}.
We minimize a cross-entropy loss $\mathcal{L}_{\mathrm{CE}}$ to ensure predicted reports are close to ground truth reports.
To improve the coherence of generated reports, we introduce an auxiliary contrastive alignment module. 
This module is designed to align the distributions of the visual and textual modalities, thereby enhancing the model's overall performance.
Initially, for each visual token embedding $\mathbf{D}_j^{(i)} \in \mathbb{R}^{S' \times F'}$, we perform a mean pooling operation along the first dimension and it results in a global representation of the entire image, denoted as $\mathbf{\tilde{D}}_j^{(i)} \in \mathbb{R}^{F'}$.
Then, we use a text encoder to encode each report $\mathbf{R}_j^{(i)}$ into a representation $\mathbf{E}_j^{(i)}$.
Subsequently, we concatenate all visual and text embeddings within the same minibatch to form a combined set:
$\{ \mathbf{\tilde{D}}_s \}_{s=1, 2, ..., N_B}$ and $\{ \mathbf{E}_s \}_{s=1, 2, ..., N_B}$, respectively.
Note that $N_B$ represents the total number of studies within the mini-batch.
The contrastive loss is defined as follows:

\begin{align}
\begin{split}
    \mathcal{L}_{\mathrm{Cont}} = \sum_{r=1}^{N_B} \frac{1}{2N_B}&(-\mathrm{log}\frac{\mathrm{exp}(\mathrm{sim}(\mathbf{\tilde{D}}_r, \mathbf{E}_r) / \tau)}{\sum_{s'=1}^{N_B}\mathrm{exp}(\mathrm{sim}(\mathbf{\tilde{D}}_r, \mathbf{E}_{s'}) / \tau)} \\
    &  -\mathrm{log}\frac{\mathrm{exp}(\mathrm{sim}(\mathbf{E}_r, \mathbf{\tilde{D}}_r)  / \tau)}{\sum_{s'=1}^{N_B}\mathrm{exp}(\mathrm{sim}(\mathbf{E}_r, \mathbf{\tilde{D}}_{s'}) / \tau)})
\end{split}
\end{align}
where $\tau$ is the temperature hyperparameter.

\para{Learning Objectives.}
Finally, our model is optimized by jointly minimizing these two objectives:
\begin{equation}
\mathcal{L} = \mathcal{L}_{\mathrm{CE}} + \lambda \cdot \mathcal{L}_{\mathrm{Cont}}.
\end{equation}
Here, $\lambda$ is a hyperparameter used to balance these two losses. Based on empirical studies, we set $\lambda$ as $1.0$. 
The ablation results of the hyperparameter $r$ can be found in the Supplementary Material.


\subsection{Curriculum Training}
\label{sec:training}

As shown in Fig.~\ref{fig:curriculum learning}, we introduce a curriculum learning strategy, unfolding in three stages to progressively enhance our model’s performance:

\begin{itemize}
\item \textbf{Stage 1: Encoder-Decoder Report Generation}: Initially, reports are generated using an encoder-decoder architecture trained on individual chest X-ray image-text pairs. This foundational step focuses solely on static data without temporal context.

\item \textbf{Stage 2: Alignment Refinement with Text Encoder}: Subsequently, a text encoder is incorporated, utilizing contrastive learning to refine the alignment between the visual and textual data. 

\item \textbf{Stage 3: Temporal Information Learning}:The final stage expands the model's capability to a longitudinal perspective.  Here, we integrate the group causal transformer to process sequences of chest X-rays, thereby incorporating temporal information into the report generation. 
\end{itemize}

These stages collectively develop a robust and comprehensive model, which is then systematically evaluated to assess its effectiveness in generating accurate and contextually relevant radiology reports.
\section{Experiments}
\label{sec:experiment}

\subsection{Experimental Setup}

\para{Dataset and Preprocessing.}
We evaluate the performance of our model on two clinical tasks: \textit{radiology report generation} and \textit{temporal medical image classification}. The used datasets are as follows:
\begin{itemize}

\item \textbf{MIMIC-CXR:} 
%
We utilize the MIMIC-CXR dataset~\cite{johnson2019mimic}, which originally comprises $377,110$ chest X-ray images and $227,835$ reports, to evaluate our model.
Aligning with previous work~\cite{chen2020generating,chen2022cross,nicolson2023improving}, we adopt the official split of the MIMIC-CXR dataset in our experiment.
However, the original dataset includes multiple lateral images, which could introduce inconsistency in longitudinal analyses.
Additionally, we observed duplicate images within the same study, bringing noise to patient-level progression analysis.
Therefore, we meticulously curated the dataset by removing lateral images and duplicates within studies for each train/validation/test set, resulting in a preprocessed dataset consisting of $145,471$ pairs for training, $1,151$ for validation, and $2,210$ for testing.
We then follow~\cite{nicolson2023improving} to preprocess images and reports.
Specifically, we resize all images to 384$\times$384 while preserving aspect ratios.
Report preprocessing involved truncating to 60 words, converting to lowercase, removing special characters, and replacing infrequent terms with placeholders.
Crucially, we organized the image-report pairs chronologically based on the ``StudyDate" metadata, preserving temporal integrity for analyzing each patient's radiological history.
Further details on dataset curation and preprocessing are available in the Supplementary Material.
Note that we re-run the publicly released code of compared methods on our curated MIMIC-CXR dataset to ensure a fair comparison.

\item \textbf{Longitudinal MIMIC-CXR:} We further devise the Longitudinal MIMIC-CXR dataset, derived from the preprocessed MIMIC-CXR-JPG dataset, to assess our model's capability in generating temporally coherent reports, following~\cite{zhu2023utilizing}.
This subset includes only patients with at least two consecutive visits.
It is worth noting that the training, validation, and test splits of the Longitudinal-MIMIC dataset correspond to the official divisions of the MIMIC-CXR dataset.
%

\item \textbf{MS-CXR-T:} 
We also assess our model's capacity for capturing temporal information using the MS-CXR-T dataset~\cite{bannur2023ms}.
This dataset consists of $1326$ multi-image frontal chest X-rays, each annotated with one of five findings.
For each finding, there are three possible states reflecting disease progression: ``Improving," ``Stable," and ``Worsening".
%

\end{itemize}

\para{Radiology Report Generation.}
We evaluate the performance of our method in radiology report generation on both the MIMIC-CXR dataset and the Longitudinal MIMIC-CXR dataset.
We compare \modelname \ with $7$ state-of-the-art (SOTA) radiology report generation models, including $\mathcal{M}^2$Transformer~\cite{cornia2020meshed}, R2Gen~\cite{chen2020generating}, R2GenCMN~\cite{chen2022cross}, $\mathcal{M}^2$TR.PROGRESSIVE~\cite{nooralahzadeh2021progressive}, XProNet~\cite{wang2022cross}, CvT-212DistilGPT2~\cite{nicolson2023improving} and DCL~\cite{li2023dynamic}.
To ensure a fair comparison, we rerun the publicly released code of these methods on our curated MIMIC-CXR dataset.
Note that $5$ SOTA models (PPKER~\cite{liu2021exploring}, ContrastiveAttention~\cite{ma2021contrastive}, 
AlignTransformer~\cite{you2021aligntransformer},
KIUT~\cite{huang2023kiut}, and METransformer~\cite{wang2023metransformer}) lack publicly available source code, thus results are cited from their original papers for reference. 
%
%
However, we note that these can't directly compared to our results due to our additional dataset preprocessing.
Additionally, the results in the RGRG~\cite{tanida2023interactive} paper, employing the Chest ImaGenome~\cite{wu2021chest} split instead of the official MIMIC-CXR split, are also for reference only but not directly comparable.
On the Longitudinal MIMIC-CXR dataset, we compare our model with both single-image based baselines, i.e., R2Gen~\cite{chen2020generating}, R2CMN~\cite{chen2022cross}, CvT-212DistilGPT2~\cite{nicolson2023improving}, etc. and longitudinal image-based baseline, i.e., Prefilling~\cite{zhu2023utilizing}.

\para{Temporal Image Classification.}
The temporal image classification task is evaluated on the MS-CXR-T dataset~\cite{bannur2023ms}.
This evaluation serves as an additional task to assess how well our model can understand and process disease progression in medical images.
We compare our approach with both temporal image-based vision language pretraining methods (\eg, BioViL-T) and single image pretraining methods (\eg, BioViL).
More information about this experiment is available in the Supplementary Material. 

\para{Evaluation Metrics.} 
In line with previous studies~\cite{pavlopoulos2022diagnostic, chen2020generating, nicolson2023improving, tanida2023interactive}, we employed a combination of Natural Language Generation (NLG) and Clinical Efficiency (CE) metrics to evaluate our report generation performance.
For NLG, we used established metrics including BLEU-n~\cite{papineni2002bleu}, which measures n-gram overlap, METEOR~\cite{banerjee2005meteor}, that accounts for recall through an $F_{\beta}$ score, ROUGE-L~\cite{lin2004rouge}, based on the longest common subsequence.
Recognizing that NLG metrics may not fully reflect clinical accuracy, we further integrated CE metrics following previous work~\cite{chen2020generating, chen2022cross, huang2023kiut, wang2023metransformer}.
Specifically, we apply CheXbert~\cite{smit2020chexbert} to label the generated reports into 14 categories (related to thoracic diseases and support devices), and then compute precision, recall, and F1 scores against ground truths.
The macro-averaged results over $14$ classes are reported, given the susceptibility of micro-averaged metrics to minor class imbalances~\cite{sorower2010literature}.
%
%
%
%
%
%
As for the temporal image classification, we predict one of ``improving", ``stable", and ``worsening" for each one of the $5$ findings: Consolidation, Pleural Effusion,  Pneumonia, Pneumothorax and Edema. 
Following BioViL-T~\cite{bannur2023learning}, we use macro-accuracy across the 5 classes to evaluate the performance.

\subsection{Implementation Details}

We set the minibatch size to $16$ for single image-text pair training and to $4$ for temporal training.
Our model training was limited to a maximum of 5 studies per patient to accommodate resource limitations.
We employed the AdamW~\cite{loshchilov2017decoupled} optimizer for model optimization.
The learning rate was adjusted according to the training stage, with detailed strategies provided in the Supplementary Materials.
The training was early stopped if the validation BLEU-4 score did not improve over 10 consecutive epochs.
All experiments were conducted using two Nvidia GeForce RTX 3090 GPUs.

\begin{table*}[t]
    \centering
    \caption{
    Natural Language Generation (NLG) and Clinical Efficacy (CE) metrics on MIMIC-CXR.
    The \textbf{Best} and \underline{second-best} results of each metric are shown in \textbf{bold} and \underline{underline}, respectively.
    $\dagger$ indicates the results are cited from their original papers.
    Since our study involves necessary data cleaning for longitudinal analysis, these results are not strictly comparable to ours.
    %
    Results without $\dagger$ were obtained by re-running publicly available code
    on the same preprocessed dataset used in our study.
    }
    \label{tab: nlg_mimic}
    \resizebox{0.95\textwidth}{!}{
    \begin{tabu}{c | c | c c c c c c | c c c}
        \toprule
        & & \multicolumn{6}{c|}{NLG} & \multicolumn{3}{c}{CE} \\
        \midrule
        Method & Year & BLEU-1 & BLEU-2 & BLEU-3 & BLEU-4 & METEOR & ROUGE-L & Precision & Recall & F1  \\
        \midrule
        $\mathcal{M}^2$Transformer~\cite{cornia2020meshed} & 2019 & $0.352$ & $0.211$ & $0.138$ & 
        $0.096$ & $0.128$ & $0.263$ & $0.239$ & $0.173$ & $0.173$ \\
        R2Gen~\cite{chen2020generating} & 2020 & $0.339$ & $0.211$ & $0.143$ & $0.103$ & $0.138$ & $0.279$ & $0.297$ & $0.189$ & $0.193$ \\
        R2GenCMN~\cite{chen2022cross} & 2021 & $0.345$ & $0.213$ & $0.143$ & $0.101$ & $0.140$ & $0.274$ & $0.354$ & $0.271$ & $0.275$ \\
        $\mathcal{M}^2$TR.PROGRESSIVE~\cite{nooralahzadeh2021progressive} & 2021 & $0.349$ & $0.204$ & $0.132$ & $0.091$ & $0.124$ & $0.255$ & $0.220$ & $0.209$ & $0.236$ \\
        XProNet~\cite{wang2022cross} & 2022 & $0.303$ & $0.188$ & $0.127$ & $0.091$ & $0.128$ & $0.268$ & $\textbf{0.419}$ & $0.230$ & $0.242$ \\
        CvT-212DistilGPT2~\cite{nicolson2023improving} & 2022 & \underline{0.372} & \underline{0.231} & \underline{0.155} & \underline{0.111} & \underline{0.149} & \underline{0.280} & \underline{$0.417$} & \underline{$0.295$} & \underline{$0.306$} \\
        DCL~\cite{li2023dynamic} & 2023 & $0.263$ & $0.153$ & $0.099$ & $0.071$ & $0.117$ & $0.211$ & $0.303$ & $0.232$ & $0.229$ \\
        \midrule
        \modelname (\textbf{Ours}) & 2024 & \textbf{0.395} & \textbf{0.248} & \textbf{0.169} & \textbf{0.122} & \textbf{0.156} & \textbf{0.285} & $0.415$ & $\textbf{0.301}$ & $\textbf{0.317}$ \\
        \midrule
        \midrule
        \multicolumn{11}{c}{
        \footnotesize{Results below are not strictly comparable due to our dataset preprocessing. For reference only.}
        } \\
        \midrule
        \rowfont{\color{gray}}
        $\mathrm{PPKED}^{\dagger}$~\cite{liu2021exploring} & 2021 & 0.360 & 0.224 & 0.149 & 0.106 & 0.149 & 0.284 & $-$ & $-$ & $-$ \\
        \rowfont{\color{gray}}
        $\mathrm{Contrastive Attention}^{\dagger}$~\cite{ma2021contrastive} & 2021 & 0.350 & 0.219 & 0.152 & 0.109 & 0.151 & 0.283 & $0.352$ & $0.298$ & $0.303$ \\
        \rowfont{\color{gray}}
        $\mathrm{AlignTransformer}^{\dagger}$~\cite{you2021aligntransformer} & 2021 & 0.378 & 0.235 & 0.156 & 0.112 & 0.158 & 0.283 & $-$ & $-$ & $-$ \\
        \rowfont{\color{gray}}
        $\mathrm{RGRG}^{\dagger}$~\cite{tanida2023interactive}& 2023 & 0.373 & 0.249 & 0.175 & 0.126 & 0.168 & 0.264 & $-$ & $-$ & $-$ \\
        \rowfont{\color{gray}}
        $\mathrm{KIUT}^{\dagger}$~\cite{huang2023kiut} & 2023 & $0.393$ & $0.243$ & $0.159$ & $0.113$ & $0.160$ & $0.285$ & $0.371$ & $0.318$ & $0.321$ \\
        \rowfont{\color{gray}}
        $\mathrm{METransformer}^{\dagger}$~\cite{wang2023metransformer} & 2023 & $0.386$ & $0.250$ & $0.169$ & $0.124$ & $0.152$ & $0.291$ & $0.364$ & $0.309$ & $0.311$ \\
        \bottomrule
    \end{tabu}
    }
\end{table*}

\subsection{Results of Radiology Report Generation}

\para{Results on MIMIC-CXR.}
Our model exhibits excellent radiology report generation capabilities, outperforming state-of-the-art models in both Natural Language Generation (NLG) and Clinical Efficiency (CE) metrics, as shown in Table~\ref{tab: nlg_mimic}. 
For NLG metrics, it notably surpasses all baseline models, notably improving over the second-best model, CvT-212DistilGPT2, by significant margins.
Specifically, compared with CvT-212DistilGPT2, our model achieves a $\Delta +5.9\%$ overall improvement on the averaged NLG metrics compared with CvT-212DistilGPT2.
%
In CE metrics, our model enhances recall and F1 by $\Delta+2.0\%$ and $\Delta+3.6\%$, respectively, compared to the second-best results.
Our precision score of $0.415$ closely approaches the best score of $0.419$. 
Additionally, we incorporate micro-based metrics for five common observations, following the methodologies of other studies~\cite{miura2020improving, tanida2023interactive}, to provide further evaluation of our method.
These results are available in the Supplementary Material.
Furthermore, our statistical analysis verifies that our model significantly outperforms the second-best approach, as detailed in Table~\ref{tab:comparison}.

\para{Results on Longitudinal MIMIC-CXR.}
Table~\ref{tab: longitudinal_mimic} presents a comparison of our model against various baseline methods in terms of Natural Language Generation (NLG) and Clinical Efficiency (CE) metrics. 
Our model outperforms both single-image and longitudinal-image-based methods in all evaluated NLG and CE metrics. 
%
%
Notably, our model achieves an increase of $\Delta+6.5\%$ on the averaged NLG metrics compared with the second-best approach CvT-212DistilGPT2.
In terms of CE metrics, our model also outperforms CvT-212DistilGPT2 in all cases, achieving the improvements of $\Delta+14.7\%$ in precision, $\Delta+12.0\%$ in recall, and $\Delta+13.0\%$ in F1 score, respectively.
Notably, our model also significantly surpasses longitudinal-image-based baseline~\cite{zhu2023utilizing},
which also utilizes prior images and reports for current report generation, 
underscoring the effectiveness of our proposed temporal data integration strategy.

\begin{table}[t!]
    \centering
    \caption{
    Results of NLG metrics (BLEU (BL), METEOR (M), ROUGE-L ($R_L$)) and CE metrics (Precision, Recall and F1) on the Longitudinal MIMIC-CXR dataset. 
    %
    Results marked with a dagger ($\dagger$) are cited from published literature Prefilling~\cite{zhu2023utilizing}. Since our curation of the Longitudinal-MIMIC dataset aligns with the approach in Prefilling~\cite{zhu2023utilizing}, these results are directly comparable to ours.
    }
    \label{tab: longitudinal_mimic}
    \resizebox{0.9\textwidth}{!}{%
    \begin{tabular}{c | c c c c c c | c c c} 
    \toprule
    & \multicolumn{6}{c|}{NLG} & \multicolumn{3}{c}{CE} \\
    \midrule
    Method & BL-1 & BL-2 & BL-3 & BL-4 & M & $R_L$ & Precision & Recall & F1 \\
    \midrule
    \multicolumn{7}{l}{\textit{Baselines based on single images}} \\
    $\mathrm{AoANet}^{\dagger}$~\cite{huang2019attention} & $0.272$ & $0.168$ & $0.112$ & $0.080$ & $0.115$ & $0.249$ & - & - & - \\
    $\mathrm{CNN+Trans}^{\dagger}$ & $0.299$ & $0.186$ & $0.124$ & $0.088$ & $0.120$ & $0.263$ & - & - & - \\
    $\mathrm{Transformer}^{\dagger}$ & $0.294$ & $0.178$ & $0.119$ & $0.085$ & $0.123$ & $0.256$ & - & - & - \\
    $\mathrm{R2Gen}^{\dagger}$~\cite{chen2020generating} & $0.302$ & $0.183$ & $0.122$ & $0.087$ & $0.124$ & $0.259$ & - & - & - \\
    $\mathrm{R2CMN}^{\dagger}$~\cite{chen2022cross} & $0.305$ & $0.184$ & $0.122$ & $0.085$ & $0.126$ & $0.265$ & - & - & - \\
    CvT-212DistilGPT2~\cite{nicolson2023improving} & \underline{$0.365$} & \underline{$0.226$} & \underline{$0.151$} & \underline{$0.107$} & \underline{$0.143$} & \underline{$0.275$} & \underline{$0.367$} & \underline{$0.258$} & \underline{$0.261$}  \\
    \midrule
    \midrule
    \multicolumn{7}{l}{\textit{Baselines based on longitudinal images}} \\
    $\mathrm{Prefilling}^{\dagger}$~\cite{zhu2023utilizing} & $0.343$ & $0.210$ & $0.140$ & $0.099$ & $0.137$ & $0.271$ & - & - & - \\
    \midrule
    \modelname (\textbf{Ours}) & $\textbf{0.389}$ & $\textbf{0.242}$ & $\textbf{0.163}$ & $\textbf{0.117}$ & $\textbf{0.155}$ & $\textbf{0.282}$ & $\textbf{0.421}$ & $\textbf{0.289}$ & $\textbf{0.295}$ \\ 
    \bottomrule
    \end{tabular}
    }
\end{table}

\subsection{Results of Temporal Image Classification}
The temporal image classification performance on MS-CXR-T is shown in Table.~\ref{tab:temporal classification}.
%
%
We divided the dataset into training, validation, and test sets with a $70\%$ / $10\%$ / $20\%$ ratio. 
In the finetuning phrase, we employ our pretrained image encoder and group causal transformer (these two modules remain frozen) to extract representations from pairs of images, and then only train a linear layer to make predictions.
It is observed that \modelname \ achieves the best performance across $4$ diseases and achieves the second-best performance on edema.
Specifically, our model improve the macro-accuracy than the second best results by $\Delta+11.1\%$, $\Delta+4.9\%$, $\Delta+14.7\%$ and $\Delta+3.7\%$ on consolidation, pleural effusion, pneumonia, pneumothorax, respectively.
These advancements further underscore the effectiveness of our proposed group causal transformer in capturing the progression of diseases and extracting semantics from longitudinal studies.

\begin{table}[t!]
    \caption{
    Temporal medical image classification performance on MS-CXR-T. Macro-accuracy ([$\%$]) are used as the metric. The \textbf{Best} and \underline{second-best} results are shown in \textbf{bold} and \underline{underline}, respectively. Note that Pl.effusion denotes pleural effusion.
    }
    \label{tab:temporal classification}
    \centering
    \resizebox{0.9\textwidth}{!}{%
    \begin{tabular}{c c c c c c c}
    \toprule
    \textbf{Method} & \textbf{Pre-train} & \textbf{Consolidation} & \textbf{Pl. effusion} & \textbf{Pneumonia} & \textbf{Pneumothorax} & \textbf{Edema} \\
    \midrule
    Random & - & 32.3 & 31.6 & 30.3 & 39.0 & 34.9 \\
    ResNet & ImageNet & 37.5 & 39.0 & 48.4 & 45.3 & 42.5 \\
    BioViL~\cite{boecking2022making} & Static & 42.9 & 41.4 & 47.9 & 42.8 & 40.7 \\
    BioViL-T~\cite{bannur2023learning} & Temporal & \underline{45.0} & \underline{46.3} & \underline{52.0} & \underline{50.1} & \textbf{52.0} \\
    \modelname (\textbf{Ours}) & Temporal & \textbf{56.1} & \textbf{51.2} & \textbf{66.7} & 
    \textbf{54.8} & \underline{48.1} \\
    \bottomrule
    \end{tabular}
    }
\end{table}

\begin{table*}[t!]
    \centering
    \caption{
    Comparison of CvT-212DistilGPT2 and \modelname \  with $95\%$ confidence intervals, which are computed using non-parametric bootstrap.
    }
    \label{tab:comparison}
    \resizebox{0.9\textwidth}{!}{
        \begin{tabular}{c c c c c c c}
        \toprule
        Method & BLEU-1 & BLEU-2 & BLEU-3 & BLEU-4 & METEOR & ROUGE-L \\
        \midrule
        \multirow{2}{*}{CvT-212DistilGPT2} & 0.372 & 0.231 & 0.155 & 0.111 & 0.280 & 0.149 \\
        & (0.367, 0.377) & (0.227, 0.235) & (0.151, 0.158) & (0.107, 0.114) & (0.276, 0.283) & (0.147, 0.151) \\
        \multirow{2}{*}{\modelname (\textbf{Ours})} & 0.396 & 0.248 & 0.168 & 0.122 & 0.285 & 0.156 \\
        & (0.392, 0.400) & (0.244, 0.252) & (0.164, 0.172) & (0.118, 0.125) & (0.281, 0.288) & (0.154, 0.157) \\
        \midrule
        \multirow{2}{*}{Gains} & \highlight{+0.023} & \highlight{+0.017} & \highlight{+0.014} & \highlight{+0.011} & \highlight{+0.005} & \highlight{+0.006} \\
        & (+0.017, +0.030) & (+0.011, +0.023) & (+0.008, +0.019) & (+0.005, +0.016) & (0.0, +0.01) & (+0.003, +0.009) \\
        \bottomrule
        \end{tabular}
    }
\end{table*}

\begin{table*}[t]
    \centering
    \caption{
     Ablation study of different components (``CL" represents the auxiliary contrastive alignment module, and ``Temporal" denotes the group causal transformer for capturing longitudinal information).
    %
    %
    On each dataset, the row 1, 2 and 4 corresponds to the Stage 1, Stage 2, and Stage 3 of our curriculum learning strategy, respectively.
    Row 3 showcases a variant trained using our temporal approach without the contrastive learning component for comparison. 
    The relative improvements in the average of all NLG metrics compared with baseline is presented in the ``AVG.$\Delta$" column.
    }
    \label{tab:ablation_nlg}
    \resizebox{0.9\textwidth}{!}{%
    \begin{tabular}{c | c c | c c c c c c c}
    \toprule
    Dataset & CL & Temporal & BLEU-1 & BLEU-2 & BLEU-3 & BLEU-4 & METEOR & ROUGE-L & AVG.$\Delta$ \\
    \midrule
    \multirow{4}{*}{MIMIC-CXR} & & & 0.372 & 0.231 & 0.155 & 0.111 & 0.149 & 0.280 & $-$ \\
    & \checkmark & & \underline{0.390} & 0.239 & 0.161 & \underline{0.117} & \underline{0.153} & 0.280 & $+3.3\%$ \\
    &  & \checkmark & $0.388$ & \underline{$0.240$} & \underline{$0.162$} & $0.116$ & \underline{$0.153$} & \underline{$0.283$} & $+3.4\%$ \\
    & \checkmark & \checkmark & \textbf{0.395} & \textbf{0.248} & \textbf{0.169} & \textbf{0.122} & \textbf{0.156} & \textbf{0.285} & $+5.9\%$ \\
    \midrule
    \multirow{4}{*}{Lon-MIMIC} & & & $0.365$ & $0.225$ & $0.151$ & $0.107$ & $0.143$ & $0.275$ & $-$ \\
    & \checkmark & & $0.375$ & $0.232$ & $0.155$ & $0.110$ & $0.146$ & $0.277$ & $+2.3\%$ \\
    & & \checkmark & \underline{$0.380$} & \underline{$0.237$} & \underline{$0.160$} & \underline{$0.115$} & \underline{$0.153$} & $\textbf{0.283}$ & $+4.9\%$ \\
    & \checkmark & \checkmark & $\textbf{0.389}$ & $\textbf{0.242}$ & $\textbf{0.163}$ & $\textbf{0.117}$ & $\textbf{0.155}$ & $\underline{0.282}$ & $+6.5\%$ \\
    \bottomrule
    \end{tabular}
    }
\end{table*}

\subsection{Ablated Analysis of Our Framework}
\para{Effect of Auxiliary Contrastive Alignment.}
We delve into the impacts of incorporating our auxiliary contrastive alignment module, as delineated in Table~\ref{tab:ablation_nlg}.
It is observed that incorporating the contrastive learning objective yields improvements in all NLG metrics compared to the baseline for both datasets, suggesting enhanced consistency in report generation. 
%
%
%
Notably, we observed a augmentation of $+3.3\%$ in average NLG metrics for the MIMIC-CXR dataset and a $+2.3\%$ improvement for the Longitudinal MIMIC-CXR dataset, underscoring the value of our contrastive alignment module in report generation.
%

\para{Effect of Temporal Aggregation Module.}
We evaluate the impact of integrating our temporal aggregation module in Table~\ref{tab:ablation_nlg}. 
It is observed that this module significantly enhances the NLG metrics across the MIMIC-CXR and Longitudinal MIMIC-CXR datasets. 
Specifically, on the Longitudinal MIMIC-CXR dataset, 
%
it achieves $+4.9\%$ improvement in the averaged NLG metrics,
compared to the baseline upon integrating this module.
When combined with a model trained using contrastive learning, 
%
the improvement on the averaged NLG metrics compared with baseline further increases to $+6.5\%$, which marks a significant enhancement than $+2.3\%$ (Row 2).
%
%
This pattern is consistent across datasets, underscoring the temporal aggregation module's effectiveness in leveraging patient histories for generating more accurate reports.

\para{Effect of Curriculum Learning Strategy.}
We evaluate the impact of our curriculum learning strategy on model performance in Table~\ref{tab:ablation_nlg}.
Our analysis reveals that, across both the MIMIC-CXR and Longitudinal MIMIC-CXR datasets, the models incorporating contrastive learning alignment consistently outperform the baseline.
Furthermore, our final model, which integrates both contrastive learning and temporal aggregation module, shows the best performance across the majority of metrics, highlighting the combined benefits of these approaches.
For a detailed comparison between joint and curriculum-based training, please refer to the Supplementary Material.

\begin{figure}[t!]
    \centering
    \begin{minipage}{0.38\textwidth}
        \includegraphics[width=\linewidth]{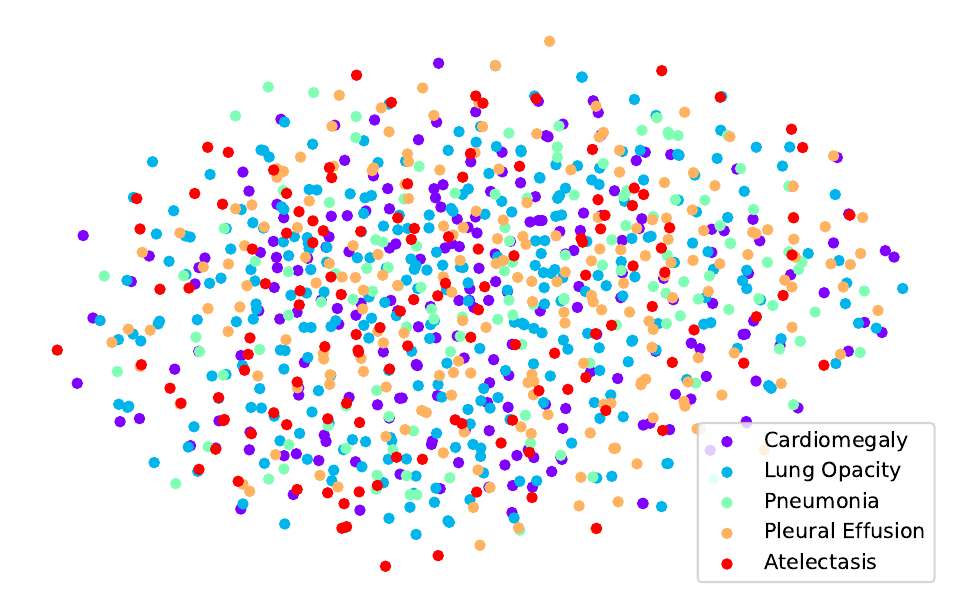}
        \centering
        (a) CvT-212DistilGPT2
    \end{minipage}
    \hfill
    \begin{minipage}{0.38\textwidth}
        \includegraphics[width=\linewidth]{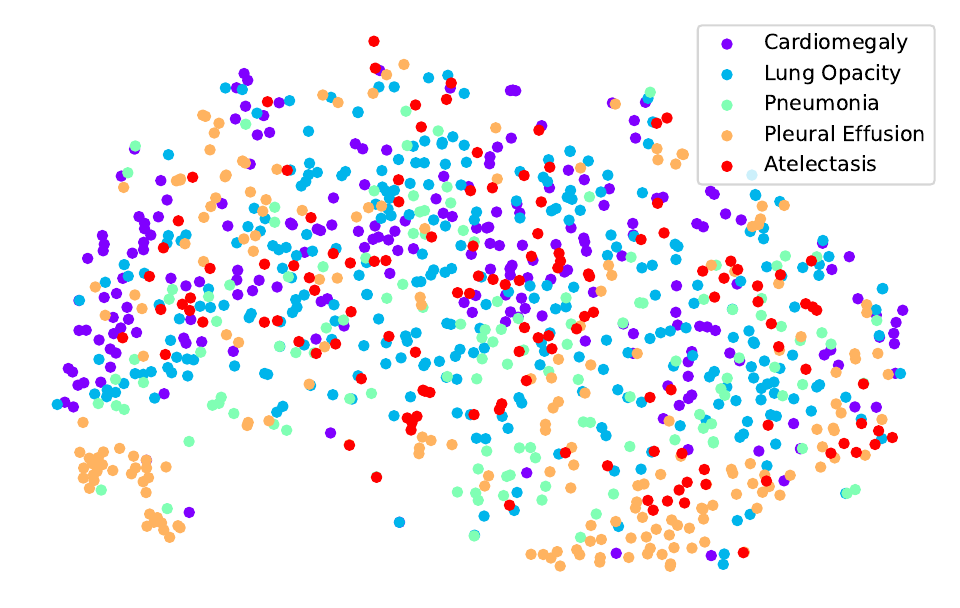}
        \centering
        (b) Our approach
    \end{minipage}
    \caption{
    Embedding visualization of MIMIC-CXR images in CvT-212DistilGPT2 and our model with t-SNE.
    }
    \label{fig:tsne-visualization}
\end{figure}

\begin{figure*}[t!]
    \centering
    \includegraphics[width=0.9\linewidth]{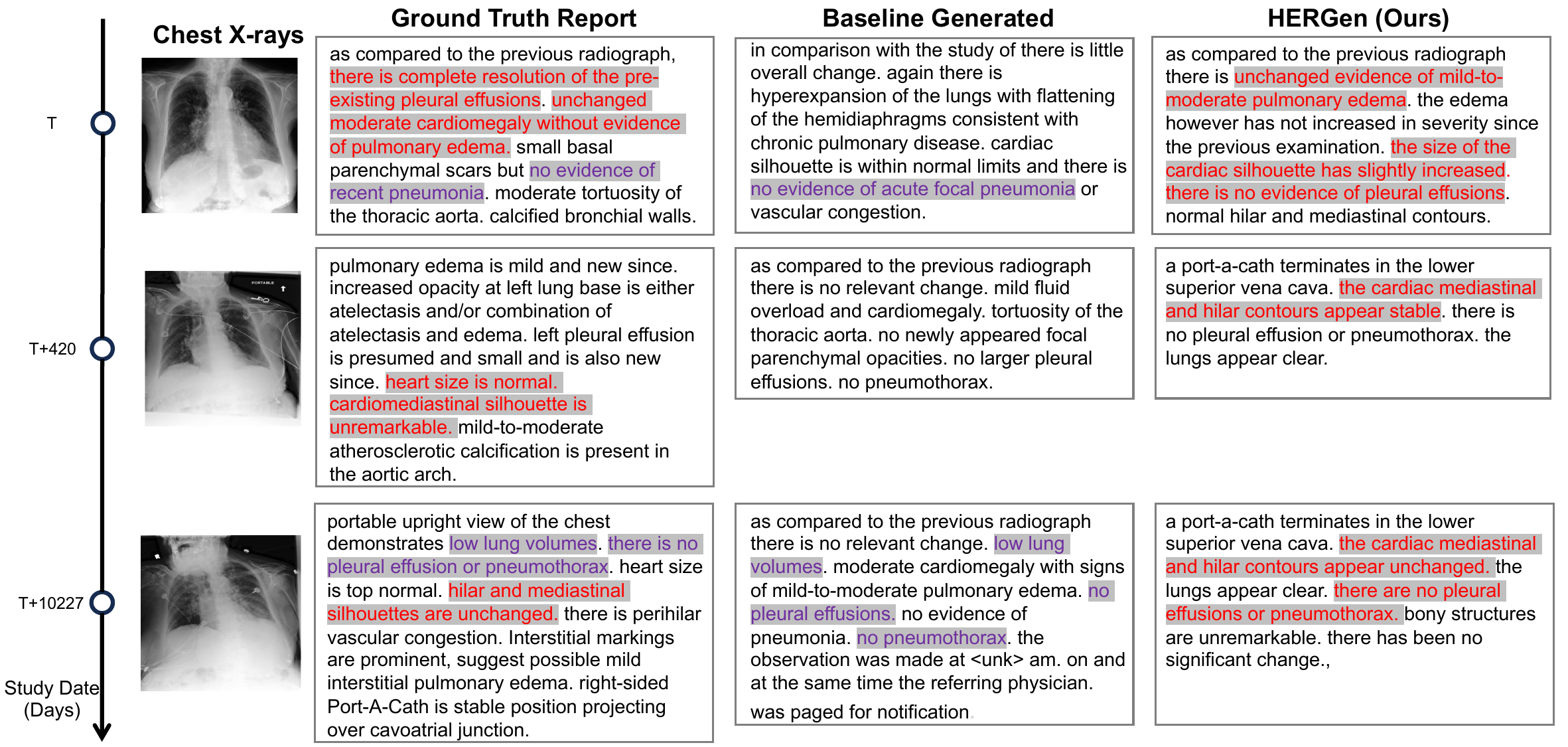}
    \caption{
    This case study compares radiology report predictions for a patient by our model and CvT-212DistilGPT2.
    Text highlighted in gray indicates words or their synonyms found in both the predicted and ground truth reports.
    Purple highlights denote similar matches in the baseline-generated (CvT-212DistilGPT2) reports and ground truth, while red highlights show similar matches in our model's reports and ground truth.
    From top to bottom, the chest X-rays are chronologically ordered.
    Here $T$ denotes the study date of the first study.}
    \label{fig:casestudy}
\end{figure*}

\subsection{Qualitative Results}

\para{Case Study of Generated Reports.}
Fig.~\ref{fig:casestudy} presents a case study comparing reports generated by our model with those from CvT-212DistilGPT2 for a given patient. 
The comparison shows that reports from our model align more clinical findings with the ground truth. 
Moreover, our model correctly generates more comparative statements, such as ``appear stable" or ``appear unchanged", suggesting its superiority to capture temporal information.
These findings underscore our model's proficiency in report generation by (1) identifying disease-specific features through consistent anatomical structures in patient-level CXRs and (2) generating time-comparative sentences.
%

\para{Visualization of Learned Embeddings.}
Fig.~\ref{fig:tsne-visualization} shows T-SNE visualization of image embeddings from a dataset constructed following $5\times 200$ MIMIC-CXR dataset proposed in MedCLIP~\cite{wang2022medclip}. 
%
%
We calculated the $50$-th percentile of the cosine similarity matrix within each class, revealing that our model achieves a higher average similarity (0.0919) compared to CvT-212DistilGPT2 (0.0245), indicating better disease-specific feature extraction.
\section{Conclusion}
In this paper, we present a novel framework to enhance radiology report generation by utilizing the varying-size patient histories.
By integrating a novel group casual transformer, our model effectively aggregates temporal information of longitudinal data.
Besides, our framework optimize an auxiliary contrastive alignment module to further align image and textual data.
Moreover, a curriculum learning strategy is employed to sequentially optimize these modules, thereby progressively improving model performance. 
Our extensive experiments demonstrate the model's capability to generate clinically precise reports and extract meaningful insights from historical data.

\para{Limitations and Future Work.}
One potential limitation of our method is that the model's alignment operates within the embedding space without accounting for anatomical consistencies in longitudinal studies.
Additionally, we plan to expand HERGen into a more comprehensive representation learning model, thereby broadening its utility across varied downstream tasks.

\section*{Acknowledgement}
This work was partially supported by the Research Grants Council of Hong Kong (27206123 and T45-401/22-N), the Hong Kong Innovation and Technology Fund (ITS/273/22), and the National Natural Science Foundation of China (No. 62201483).

%
%
\bibliographystyle{splncs04}
\bibliography{main}

\begin{thebibliography}{10}
\providecommand{\url}[1]{\texttt{#1}}
\providecommand{\urlprefix}{URL }
\providecommand{\doi}[1]{https://doi.org/#1}

\bibitem{alfarghaly2021automated}
Alfarghaly, O., Khaled, R., Elkorany, A., Helal, M., Fahmy, A.: Automated radiology report generation using conditioned transformers. Informatics in Medicine Unlocked  \textbf{24},  100557 (2021)

\bibitem{alsentzer2019publicly}
Alsentzer, E., Murphy, J.R., Boag, W., Weng, W.H., Jin, D., Naumann, T., McDermott, M.: Publicly available clinical bert embeddings. arXiv preprint arXiv:1904.03323  (2019)

\bibitem{banerjee2005meteor}
Banerjee, S., Lavie, A.: Meteor: An automatic metric for mt evaluation with improved correlation with human judgments. In: Proceedings of the acl workshop on intrinsic and extrinsic evaluation measures for machine translation and/or summarization. pp. 65--72 (2005)

\bibitem{bannur2023ms}
Bannur, S., Hyland, S., Liu, Q., P{\'e}rez-Garc{\'\i}a, F., Ilse, M., de~Castro, D.C., Boecking, B., Sharma, H., Bouzid, K., Schwaighofer, A., et~al.: Ms-cxr-t: Learning to exploit temporal structure for biomedical vision-language processing (2023)

\bibitem{bannur2023learning}
Bannur, S., Hyland, S., Liu, Q., Perez-Garcia, F., Ilse, M., Castro, D.C., Boecking, B., Sharma, H., Bouzid, K., Thieme, A., et~al.: Learning to exploit temporal structure for biomedical vision-language processing. In: Proceedings of the IEEE/CVF Conference on Computer Vision and Pattern Recognition. pp. 15016--15027 (2023)

\bibitem{beltagy2019scibert}
Beltagy, I., Lo, K., Cohan, A.: Scibert: A pretrained language model for scientific text. arXiv preprint arXiv:1903.10676  (2019)

\bibitem{boecking2022making}
Boecking, B., Usuyama, N., Bannur, S., Castro, D.C., Schwaighofer, A., Hyland, S., Wetscherek, M., Naumann, T., Nori, A., Alvarez-Valle, J., et~al.: Making the most of text semantics to improve biomedical vision--language processing. In: European conference on computer vision. pp. 1--21. Springer (2022)

\bibitem{cao2023current}
Cao, D.J., Hurrell, C., Patlas, M.N.: Current status of burnout in canadian radiology. Canadian Association of Radiologists Journal  \textbf{74}(1),  37--43 (2023)

\bibitem{chen2022cross}
Chen, Z., Shen, Y., Song, Y., Wan, X.: Cross-modal memory networks for radiology report generation. arXiv preprint arXiv:2204.13258  (2022)

\bibitem{chen2020generating}
Chen, Z., Song, Y., Chang, T.H., Wan, X.: Generating radiology reports via memory-driven transformer. arXiv preprint arXiv:2010.16056  (2020)

\bibitem{cornia2020meshed}
Cornia, M., Stefanini, M., Baraldi, L., Cucchiara, R.: Meshed-memory transformer for image captioning. In: Proceedings of the IEEE/CVF conference on computer vision and pattern recognition. pp. 10578--10587 (2020)

\bibitem{devlin2018bert}
Devlin, J., Chang, M.W., Lee, K., Toutanova, K.: Bert: Pre-training of deep bidirectional transformers for language understanding. arXiv preprint arXiv:1810.04805  (2018)

\bibitem{gu2021domain}
Gu, Y., Tinn, R., Cheng, H., Lucas, M., Usuyama, N., Liu, X., Naumann, T., Gao, J., Poon, H.: Domain-specific language model pretraining for biomedical natural language processing. ACM Transactions on Computing for Healthcare (HEALTH)  \textbf{3}(1),  1--23 (2021)

\bibitem{huang2019attention}
Huang, L., Wang, W., Chen, J., Wei, X.Y.: Attention on attention for image captioning. In: Proceedings of the IEEE/CVF international conference on computer vision. pp. 4634--4643 (2019)

\bibitem{huang2021gloria}
Huang, S.C., Shen, L., Lungren, M.P., Yeung, S.: Gloria: A multimodal global-local representation learning framework for label-efficient medical image recognition. In: Proceedings of the IEEE/CVF International Conference on Computer Vision. pp. 3942--3951 (2021)

\bibitem{huang2023kiut}
Huang, Z., Zhang, X., Zhang, S.: Kiut: Knowledge-injected u-transformer for radiology report generation. In: Proceedings of the IEEE/CVF Conference on Computer Vision and Pattern Recognition. pp. 19809--19818 (2023)

\bibitem{jing2020show}
Jing, B., Wang, Z., Xing, E.: Show, describe and conclude: On exploiting the structure information of chest x-ray reports. arXiv preprint arXiv:2004.12274  (2020)

\bibitem{jing2017automatic}
Jing, B., Xie, P., Xing, E.: On the automatic generation of medical imaging reports. arXiv preprint arXiv:1711.08195  (2017)

\bibitem{johnson2019mimic}
Johnson, A., Lungren, M., Peng, Y., Lu, Z., Mark, R., Berkowitz, S., Horng, S.: Mimic-cxr-jpg-chest radiographs with structured labels. PhysioNet  (2019)

\bibitem{karwande2022chexrelnet}
Karwande, G., Mbakwe, A.B., Wu, J.T., Celi, L.A., Moradi, M., Lourentzou, I.: Chexrelnet: An anatomy-aware model for tracking longitudinal relationships between chest x-rays. In: International Conference on Medical Image Computing and Computer-Assisted Intervention. pp. 581--591. Springer (2022)

\bibitem{li2023dynamic}
Li, M., Lin, B., Chen, Z., Lin, H., Liang, X., Chang, X.: Dynamic graph enhanced contrastive learning for chest x-ray report generation. In: Proceedings of the IEEE/CVF Conference on Computer Vision and Pattern Recognition. pp. 3334--3343 (2023)

\bibitem{lin2004rouge}
Lin, C.Y.: Rouge: A package for automatic evaluation of summaries. In: Text summarization branches out. pp. 74--81 (2004)

\bibitem{liu2021exploring}
Liu, F., Wu, X., Ge, S., Fan, W., Zou, Y.: Exploring and distilling posterior and prior knowledge for radiology report generation. In: Proceedings of the IEEE/CVF conference on computer vision and pattern recognition. pp. 13753--13762 (2021)

\bibitem{loshchilov2017decoupled}
Loshchilov, I., Hutter, F.: Decoupled weight decay regularization. arXiv preprint arXiv:1711.05101  (2017)

\bibitem{ma2021contrastive}
Ma, X., Liu, F., Yin, C., Wu, X., Ge, S., Zou, Y., Zhang, P., Sun, X.: Contrastive attention for automatic chest x-ray report generation. arXiv preprint arXiv:2106.06965  (2021)

\bibitem{miura2020improving}
Miura, Y., Zhang, Y., Tsai, E.B., Langlotz, C.P., Jurafsky, D.: Improving factual completeness and consistency of image-to-text radiology report generation. arXiv preprint arXiv:2010.10042  (2020)

\bibitem{nicolson2023improving}
Nicolson, A., Dowling, J., Koopman, B.: Improving chest x-ray report generation by leveraging warm starting. Artificial Intelligence in Medicine  \textbf{144},  102633 (2023)

\bibitem{nooralahzadeh2021progressive}
Nooralahzadeh, F., Gonzalez, N.P., Frauenfelder, T., Fujimoto, K., Krauthammer, M.: Progressive transformer-based generation of radiology reports. arXiv preprint arXiv:2102.09777  (2021)

\bibitem{oord2018representation}
Oord, A.v.d., Li, Y., Vinyals, O.: Representation learning with contrastive predictive coding. arXiv preprint arXiv:1807.03748  (2018)

\bibitem{papineni2002bleu}
Papineni, K., Roukos, S., Ward, T., Zhu, W.J.: Bleu: a method for automatic evaluation of machine translation. In: Proceedings of the 40th annual meeting of the Association for Computational Linguistics. pp. 311--318 (2002)

\bibitem{pavlopoulos2022diagnostic}
Pavlopoulos, J., Kougia, V., Androutsopoulos, I., Papamichail, D.: Diagnostic captioning: a survey. Knowledge and Information Systems  \textbf{64}(7),  1691--1722 (2022)

\bibitem{radford2021learning}
Radford, A., Kim, J.W., Hallacy, C., Ramesh, A., Goh, G., Agarwal, S., Sastry, G., Askell, A., Mishkin, P., Clark, J., et~al.: Learning transferable visual models from natural language supervision. In: International conference on machine learning. pp. 8748--8763. PMLR (2021)

\bibitem{radford2019language}
Radford, A., Wu, J., Child, R., Luan, D., Amodei, D., Sutskever, I., et~al.: Language models are unsupervised multitask learners. OpenAI blog  \textbf{1}(8), ~9 (2019)

\bibitem{ramesh2022improving}
Ramesh, V., Chi, N.A., Rajpurkar, P.: Improving radiology report generation systems by removing hallucinated references to non-existent priors. In: Machine Learning for Health. pp. 456--473. PMLR (2022)

\bibitem{raoof2012interpretation}
Raoof, S., Feigin, D., Sung, A., Raoof, S., Irugulpati, L., Rosenow~III, E.C.: Interpretation of plain chest roentgenogram. Chest  \textbf{141}(2),  545--558 (2012)

\bibitem{ren2015faster}
Ren, S., He, K., Girshick, R., Sun, J.: Faster r-cnn: Towards real-time object detection with region proposal networks. Advances in neural information processing systems  \textbf{28} (2015)

\bibitem{rimmer2017radiologist}
Rimmer, A.: Radiologist shortage leaves patient care at risk, warns royal college. BMJ: British Medical Journal (Online)  \textbf{359} (2017)

\bibitem{sanh2019distilbert}
Sanh, V., Debut, L., Chaumond, J., Wolf, T.: Distilbert, a distilled version of bert: smaller, faster, cheaper and lighter. arXiv preprint arXiv:1910.01108  (2019)

\bibitem{serra2023controllable}
Serra, F.D., Wang, C., Deligianni, F., Dalton, J., O'Neil, A.Q.: Controllable chest x-ray report generation from longitudinal representations. arXiv preprint arXiv:2310.05881  (2023)

\bibitem{smit2020chexbert}
Smit, A., Jain, S., Rajpurkar, P., Pareek, A., Ng, A.Y., Lungren, M.P.: Chexbert: combining automatic labelers and expert annotations for accurate radiology report labeling using bert. arXiv preprint arXiv:2004.09167  (2020)

\bibitem{sorower2010literature}
Sorower, M.S.: A literature survey on algorithms for multi-label learning. Oregon State University, Corvallis  \textbf{18}(1), ~25 (2010)

\bibitem{tanida2023interactive}
Tanida, T., M{\"u}ller, P., Kaissis, G., Rueckert, D.: Interactive and explainable region-guided radiology report generation. In: Proceedings of the IEEE/CVF Conference on Computer Vision and Pattern Recognition. pp. 7433--7442 (2023)

\bibitem{thrall2018artificial}
Thrall, J.H., Li, X., Li, Q., Cruz, C., Do, S., Dreyer, K., Brink, J.: Artificial intelligence and machine learning in radiology: opportunities, challenges, pitfalls, and criteria for success. Journal of the American College of Radiology  \textbf{15}(3),  504--508 (2018)

\bibitem{vaswani2017attention}
Vaswani, A., Shazeer, N., Parmar, N., Uszkoreit, J., Jones, L., Gomez, A.N., Kaiser, {\L}., Polosukhin, I.: Attention is all you need. Advances in neural information processing systems  \textbf{30} (2017)

\bibitem{velivckovic2017graph}
Veli{\v{c}}kovi{\'c}, P., Cucurull, G., Casanova, A., Romero, A., Lio, P., Bengio, Y.: Graph attention networks. arXiv preprint arXiv:1710.10903  (2017)

\bibitem{vinyals2015show}
Vinyals, O., Toshev, A., Bengio, S., Erhan, D.: Show and tell: A neural image caption generator. In: Proceedings of the IEEE conference on computer vision and pattern recognition. pp. 3156--3164 (2015)

\bibitem{wang2022multi}
Wang, F., Zhou, Y., Wang, S., Vardhanabhuti, V., Yu, L.: Multi-granularity cross-modal alignment for generalized medical visual representation learning. Advances in Neural Information Processing Systems  \textbf{35},  33536--33549 (2022)

\bibitem{wang2022cross}
Wang, J., Bhalerao, A., He, Y.: Cross-modal prototype driven network for radiology report generation. In: European Conference on Computer Vision. pp. 563--579. Springer (2022)

\bibitem{wang2018tienet}
Wang, X., Peng, Y., Lu, L., Lu, Z., Summers, R.M.: Tienet: Text-image embedding network for common thorax disease classification and reporting in chest x-rays. In: Proceedings of the IEEE conference on computer vision and pattern recognition. pp. 9049--9058 (2018)

\bibitem{wang2023metransformer}
Wang, Z., Liu, L., Wang, L., Zhou, L.: Metransformer: Radiology report generation by transformer with multiple learnable expert tokens. In: Proceedings of the IEEE/CVF Conference on Computer Vision and Pattern Recognition. pp. 11558--11567 (2023)

\bibitem{wang2022medclip}
Wang, Z., Wu, Z., Agarwal, D., Sun, J.: Medclip: Contrastive learning from unpaired medical images and text. arXiv preprint arXiv:2210.10163  (2022)

\bibitem{wu2021cvt}
Wu, H., Xiao, B., Codella, N., Liu, M., Dai, X., Yuan, L., Zhang, L.: Cvt: Introducing convolutions to vision transformers. In: Proceedings of the IEEE/CVF international conference on computer vision. pp. 22--31 (2021)

\bibitem{wu2021chest}
Wu, J.T., Agu, N.N., Lourentzou, I., Sharma, A., Paguio, J.A., Yao, J.S., Dee, E.C., Mitchell, W., Kashyap, S., Giovannini, A., et~al.: Chest imagenome dataset for clinical reasoning. arXiv preprint arXiv:2108.00316  (2021)

\bibitem{xu2015show}
Xu, K., Ba, J., Kiros, R., Cho, K., Courville, A., Salakhudinov, R., Zemel, R., Bengio, Y.: Show, attend and tell: Neural image caption generation with visual attention. In: International conference on machine learning. pp. 2048--2057. PMLR (2015)

\bibitem{you2021aligntransformer}
You, D., Liu, F., Ge, S., Xie, X., Zhang, J., Wu, X.: Aligntransformer: Hierarchical alignment of visual regions and disease tags for medical report generation. In: Medical Image Computing and Computer Assisted Intervention--MICCAI 2021: 24th International Conference, Strasbourg, France, September 27--October 1, 2021, Proceedings, Part III 24. pp. 72--82. Springer (2021)

\bibitem{you2016image}
You, Q., Jin, H., Wang, Z., Fang, C., Luo, J.: Image captioning with semantic attention. In: Proceedings of the IEEE conference on computer vision and pattern recognition. pp. 4651--4659 (2016)

\bibitem{zhang2020radiology}
Zhang, Y., Wang, X., Xu, Z., Yu, Q., Yuille, A., Xu, D.: When radiology report generation meets knowledge graph. In: Proceedings of the AAAI Conference on Artificial Intelligence. vol.~34, pp. 12910--12917 (2020)

\bibitem{zhang2022contrastive}
Zhang, Y., Jiang, H., Miura, Y., Manning, C.D., Langlotz, C.P.: Contrastive learning of medical visual representations from paired images and text. In: Machine Learning for Healthcare Conference. pp. 2--25. PMLR (2022)

\bibitem{zhu2023utilizing}
Zhu, Q., Mathai, T.S., Mukherjee, P., Peng, Y., Summers, R.M., Lu, Z.: Utilizing longitudinal chest x-rays and reports to pre-fill radiology reports. arXiv preprint arXiv:2306.08749  (2023)

\end{thebibliography}
\end{document}